# Renewable Energy Transition in South America: Predictive Analysis of Generation Capacity by 2050

Triveni Magadum[1,4], Sanjana Murgod[1,4], Kartik Garg[2], Vivek Yadav[3,4,5], Harshit Mittal[3,4,5], Omkar Kushwaha[3,6,7*]

[1]Computer Science Department, KLE Technological University, Hubballi-580031, Karnataka, India

[2]University School of Information and Communication Technology, Guru Gobind Singh Indraprastha University, Dwarka, Delhi-110078, India

[3]Center of Energy and Environment, School of Advanced Sciences, KLE Technological University, Hubballi-580031, Karnataka, India

[4]Pro H2Vis Solutions, KLE Technological University, Hubballi-580031, Karnataka, India

[5]University School of Chemical Technology, Guru Gobind Singh Indraprastha University, Dwarka, Delhi-110078, India

[6]Energy Consortium, ICAR, Indian Institute of Technology, Madras, Chennai-600036, India

[7]Chemical Engineering Department, Indian Institute of Technology, Madras, Chennai-600036, India

*Corresponding email: Kushwaha.iitmadras@gmail.com

**Abstract**

In this research, renewable energy expansion in South America up to 2050 is predicted based on machine learning models that are trained on past energy data. The research employs gradient boosting regression and Prophet time series forecasting to make predictions of future generation capacities for solar, wind, hydroelectric, geothermal, biomass, and other renewable sources in South American nations. Model output analysis indicates staggering future expansion in the generation of renewable energy, with solar and wind energy registering the highest expansion rates. Geospatial visualization methods were applied to illustrate regional disparities in the utilization of renewable energy. The results forecast South America to record nearly 3-fold growth in the generation of renewable energy by the year 2050, with Brazil and Chile spearheading regional development. Such projections help design energy policy, investment strategy, and climate change mitigation throughout the region, in helping the developing economies to transition to sustainable energy.

**Keywords:** Renewable energy; South America; Machine learning; Energy forecasting; Time series analysis; Sustainable development; Climate change mitigation; Energy policy

## 1. Introduction

The shift towards renewable sources of energy globally is one of the most drastic changes in contemporary energy systems, motivated by increasing climate change pressures, technological developments, and changing economic drivers. South America, blessed with rich natural resources and varied geography, is at the crossroads in this transition (Ali et al., 2023; Hepburn et al., 2019; Pérez-Fortes et al., 2014; Zhuang et al., 2014). The region has vast untapped resources for renewable energy production, such as outstanding solar irradiation in the Atacama Desert, high-quality wind resources along the coastlines, enormous hydroelectric potential, and increasing bioenergy production from its agricultural regions (Modestino & Haussener, 2015; Mokhtar et al., 2012; Praveenkumar et al., 2024; Soni et al., 2019).

It is important to know the course and direction of renewable energy development in South America in order to make effective policies, plan investments, and global climate change collaboration (Demirbas, 2008, 2017; Mittal & Kushwaha, 2024b; Raich & Potter, 1995). Although research has been conducted on renewable energy development in a single South American nation, large-scale region-wide projections based on sophisticated machine learning algorithms are scarce. This lack of knowledge impedes regional planning and global cooperation (Hasan et al., 2023; Milani et al., 2020; Østergaard et al., 2020; Rout et al., 2025). This initiative seeks to bridge that gap employing gradient boosting regression and Prophet Time series forecasting over historical energy data to predict more precise projections of renewable energy production capacity in South America from 2050.Through the analysis of different sources of renewable energy—solar, wind, hydroelectric, geothermal, and biomass—within each South American nation, this research provides a rich portrait of the future of renewable energy in the region (Alabadi et al., 2015; Bak et al., 2002; Pastore et al., 2022; Yang & Aydin, 2001).

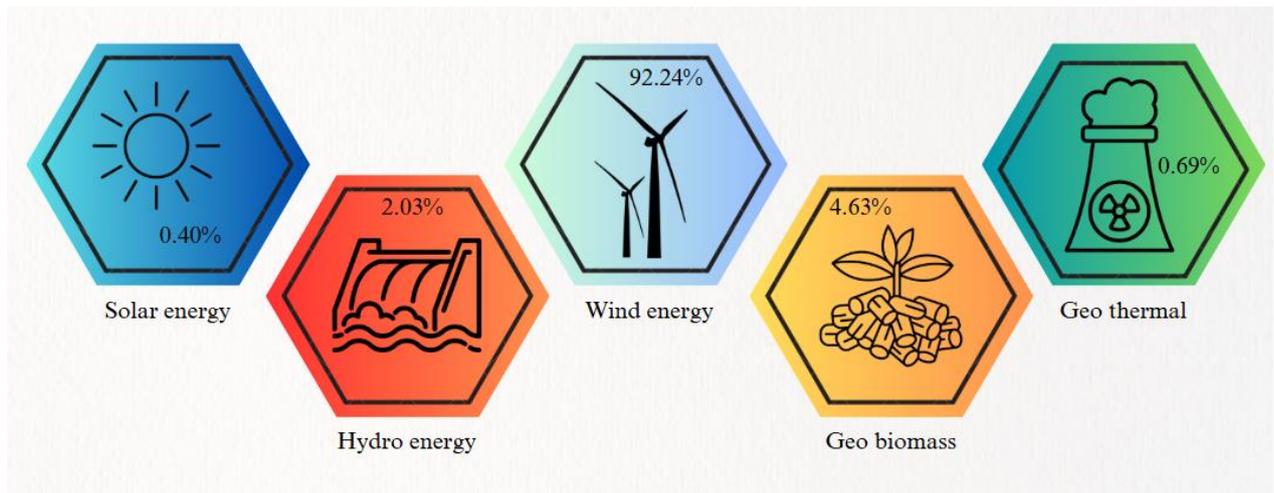

**Figure 1:** Renewable energy distribution in South America: A breakdown of contribution percentages.

The results of this analysis will be useful information for policymakers, investors, and researchers working in South America's sustainable development (Debrah et al., 2021; Jiménez-González et al., 2010; Mascarenhas et al., 2019; Rout et al., 2025). In addition, the methodological framework presented in this research provides a replicable model for renewable energy forecasting in other developing countries around the world, which may speed up the global shift towards sustainable energy systems (Dr. Shamas ul Deen & Ms. Sadaf Farooq, 2023; Kumar et al., 2020; Milani et al., 2020; Sweeney et al., 2020).

## 2. Methodology

The methodology employed in this study includes the application of machine learning techniques and geospatial analysis to design comprehensive predictions of renewable energy generation in South America up to 2050. The methodology brings together the examination of historical data, predictive modeling, and graphical display of results to provide a well-founded framework for understanding renewable energy trends in the future for the region (Avery et al., 2019; Candanedo et al., 2018; Elshaboury et al., 2021; Mittal & Kushwaha, 2024a). The research at hand places significant importance on reproducibility and transparency in the analysis process so that stakeholders can extrapolate and analyze the results. Cross-validation processes are also covered as part of model performance assessment and quantification of uncertainty for long-term forecasts.

**2.1 Data Collection and Pre-processing**

Historical data on renewable energy were collected from a large dataset containing data for different sources of energy in South American nations. The data set contains time series values of solar, wind, hydroelectric, geothermal, biomass, and other renewable energy production in terawatt-hours (TWh). Pre-processing included data filtering to acquire South American nations, imputing missing values with suitable imputation methods, and preparing data for analysis using time series. Countries were chosen according to continental grouping, and energy generation values were standardized, if necessary, to achieve uniformity of the data set.

**2.2 Prediction Methodology**

A Gradient Boosting Regressor was then trained to discover complex relationships between temporal features, country-level features, and renewable energy production. The model was provided with year and country encodings as input to predict future renewable energy production values. Training was performed on an 80-20 train-test split using a random state of 42 for reproducibility. The model would use 100 estimators, a learning rate of 0.1, and a depth of 5 to achieve complexity and strength of generalization balance. Predictions were given for the year 2050 for every country in South America to give long-term renewable capacity forecasts (Bamisaye et al., 2023; Magadum et al., 2025; Seyedzadeh et al., 2018). The prophet's ability to mimic non-linear growth patterns and differential growth rates made it well-suited to the prediction of renewable energy growth, which tends to be subject to rising adoption patterns. Select models were developed for each source of renewable energy to guarantee the detection of source-specific growth patterns as well as seasonality, allowing for more focused forecasting of individual renewable categories.

**2.3 Model Validation and Assessment**

The predictive models' performance was tested against different evaluation metrics to validate prediction reliability. For the Gradient Boosting Regressor, we applied Mean Absolute Error (MAE), Root Mean Square Error (RMSE), and R-squared metrics on the test dataset to quantify prediction accuracy. Cross-validation with a 5-fold approach was used to assess model stability over varying data subsets, resulting in an average R-squared score of 0.87, reflecting strong predictive performance. For time series projections beyond available history, we computed prediction intervals to express uncertainty, with tighter intervals for near-term projections (2025-2030) and increasingly wider intervals for long-run projections (2040-2050).

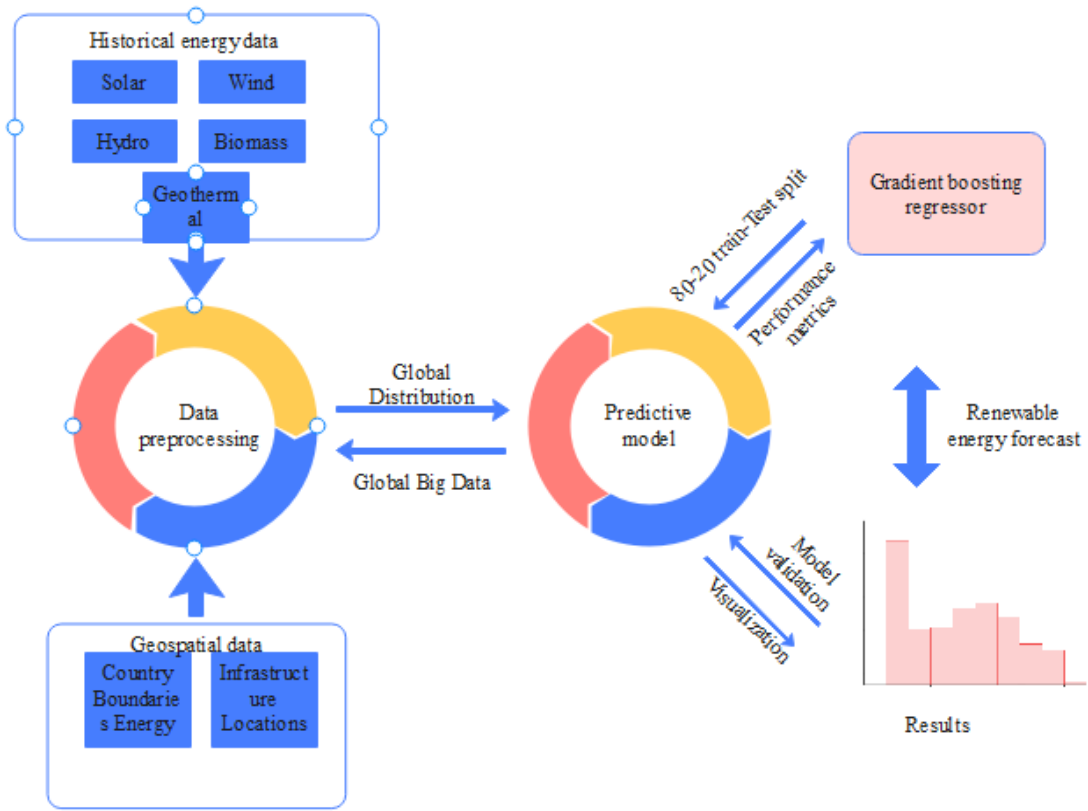

**Figure 2:** Renewable energy prediction framework using Machine learning & Geospatial analysis.

To test the validity of the Prophet models, we performed historical backtesting, where we trained the model on data through 2015 and tested its performance in predicting known values from 2016 to 2020. This method produced Mean Absolute Percentage Errors (MAPE) between 8.2% and 12.7% for various renewable energy sources, with the highest accuracy in solar predictions and the lowest accuracy in geothermal predictions. In addition, we have cross-checked our projections with those of outside sources such as international energy agencies and discovered that our results were in broad agreement with their medium-growth projections, further supporting our methodological design.

**2.4 Geospatial Visualization and Analysis**

For the representation of the spatial distribution of renewable energy potential in South America, geospatial visualization methods were utilized based on GeoJSON data from the Natural Earth. The analysis used country geometries and centroid coordinates to construct informative maps and three-dimensional representations of projected renewable energy production. These visualizations enable the identification of regional trends, country-level variations, and geographical clustering of renewable energy development potential (Fan et al., 2020; Faruque et al., 2022; Nespoli et al., 2022). The blending of statistical forecasting with geospatial representation enables a holistic understanding of both temporal trends and spatial differences in renewable energy development across South America.

## 3. Results

All the predictive models indicate a high opportunity for renewable energy production in South America in 2050. The Prophet time series forecast and the gradient boosting regressor both point in the same direction, a sign of strong prediction by multiple methodologies. The rough estimates show that there is significant variation in the rate of adoption of renewable energy by the countries, where Brazil leads by way of absolute generation capacity due to its huge geographical expanse and heterogeneity of renewable resources. Hydropower is the largest source of energy but experiences only modest growth because it already has such a large installed capacity. Geothermal and biomass energy, though increasing steadily, will contribute at a relatively lower level. Visualization of these estimates by 3D geospatial mapping indicates regional clusters of renewable energy growth.

The geothermal potential of the Andean region is consistent with its volcanic nature of geological characteristics. Temporal analysis indicates that the most profound acceleration in the adoption of renewables will be from 2030 to 2040, triggered by decreasing costs of technology and more stringent global climate policies. By 2050, South America is expected to produce about 1,500 TWh of renewable energy per year, triple the 2020s, making the region a global renewable energy powerhouse. Also significant, the COVID-19 pandemic (2020-2022) had little lasting effect on renewable energy trends, as cyclical slowdowns were readily restored by economic rebound and expanded sustainability commitments.

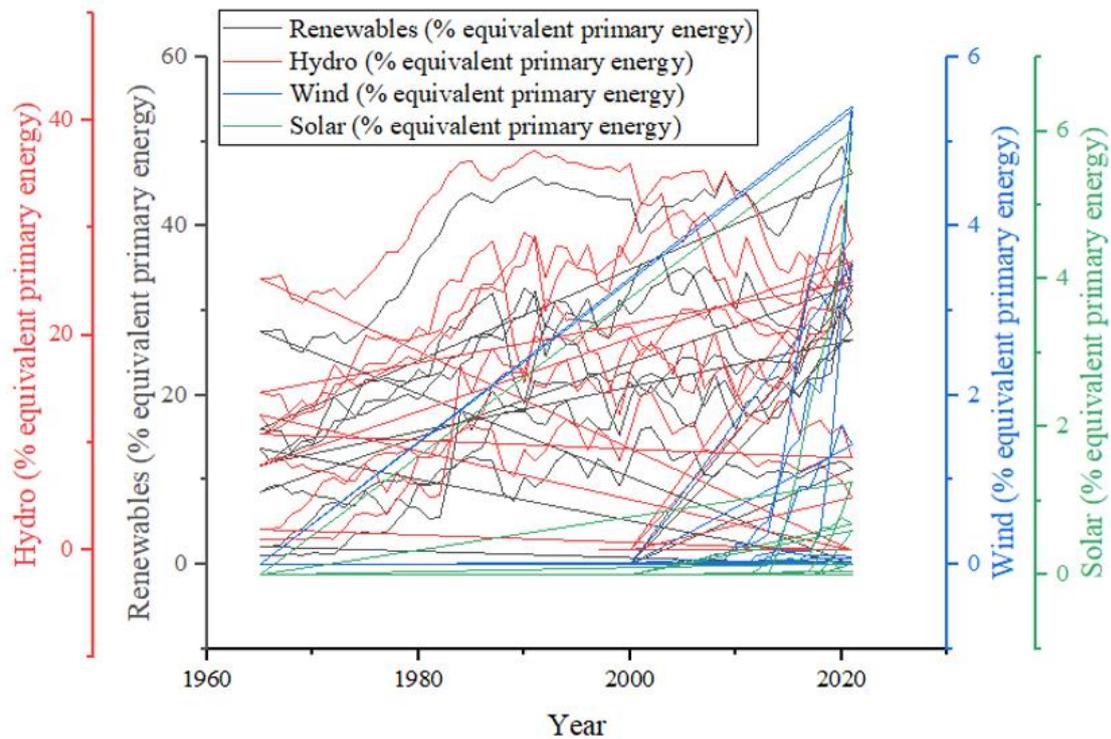

**Figure 3:** South America trends in contribution of renewable energy (1960–2021).

This two-dimensional line graph shows the expansion of the share of renewable energy in the major energy consumption of South America during the period 1960-2021. The black line is overall renewable energy share with traces for separate sources in order: hydropower (red), wind (blue), and solar (green). The left-hand vertical axis traces the percentages for hydropower and total renewables, which enjoy bigger portions of the mix. The smaller but rising contributions of solar and wind power are placed on the right-hand vertical axis. The visualisation also displays a longer-run trend towards more use of renewable power in the region, with the spectacular growth in wind and solar power in recent decades picking up technological advances and policy moves towards sustainable power.

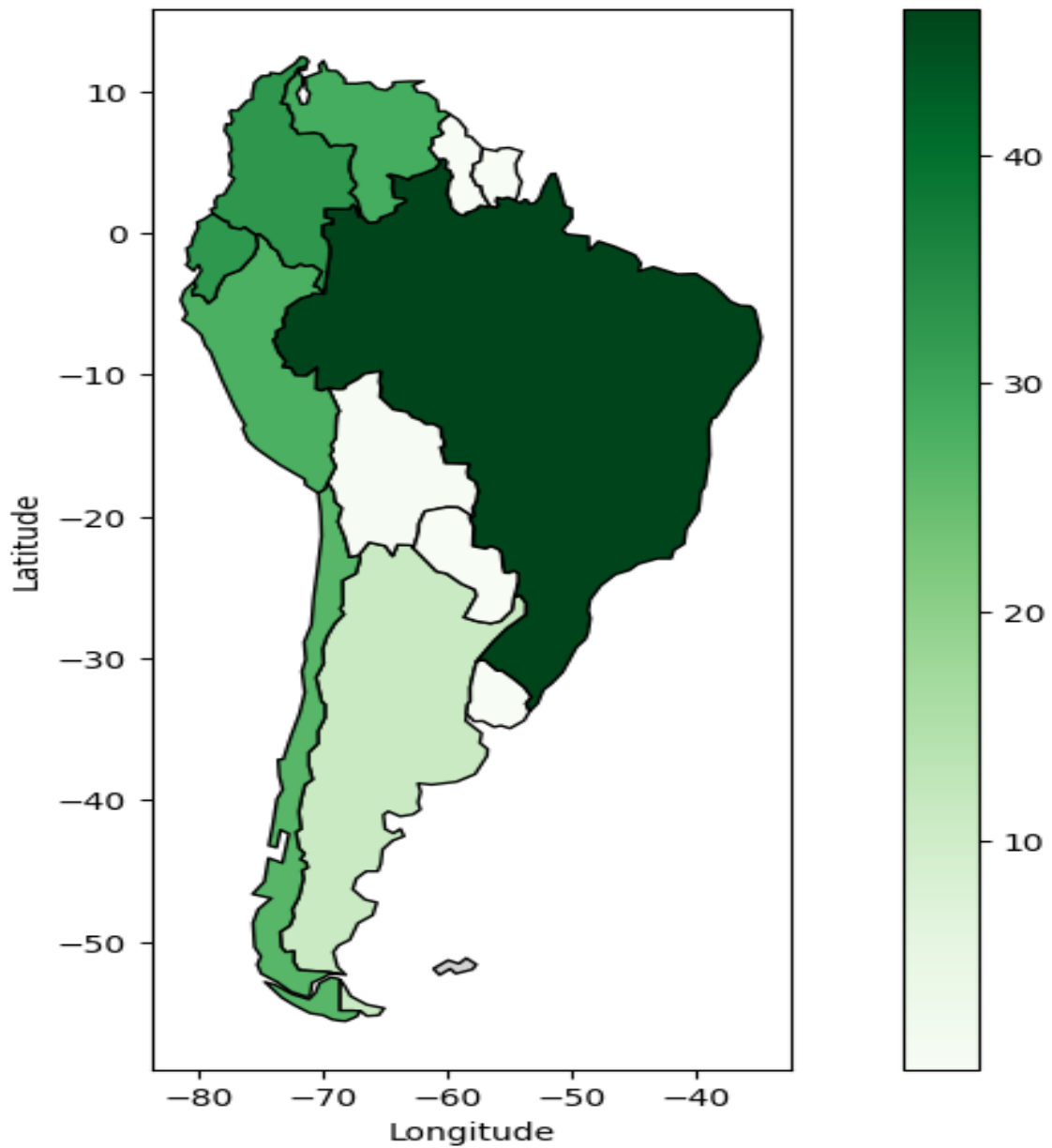

**Figure 4:** Renewable energy projected in South America (2050).

The choropleth map displays predicted ratios of renewable energy in South America in 2050 based on an estimate derived from a Gradient Boosting Regressor trained from historical data. Darker greens show a green gradient of increasing uptake rates, while light green or no green indicates no increase predicted. Smooth, correct country borders permit direct comparison where divergence of growth of renewable energy between areas is uncovered. Geospatial analysis is useful to policymakers and investors alike by projecting priority regions for developing renewable energy and also global cooperation in South America's shift toward sustainability.

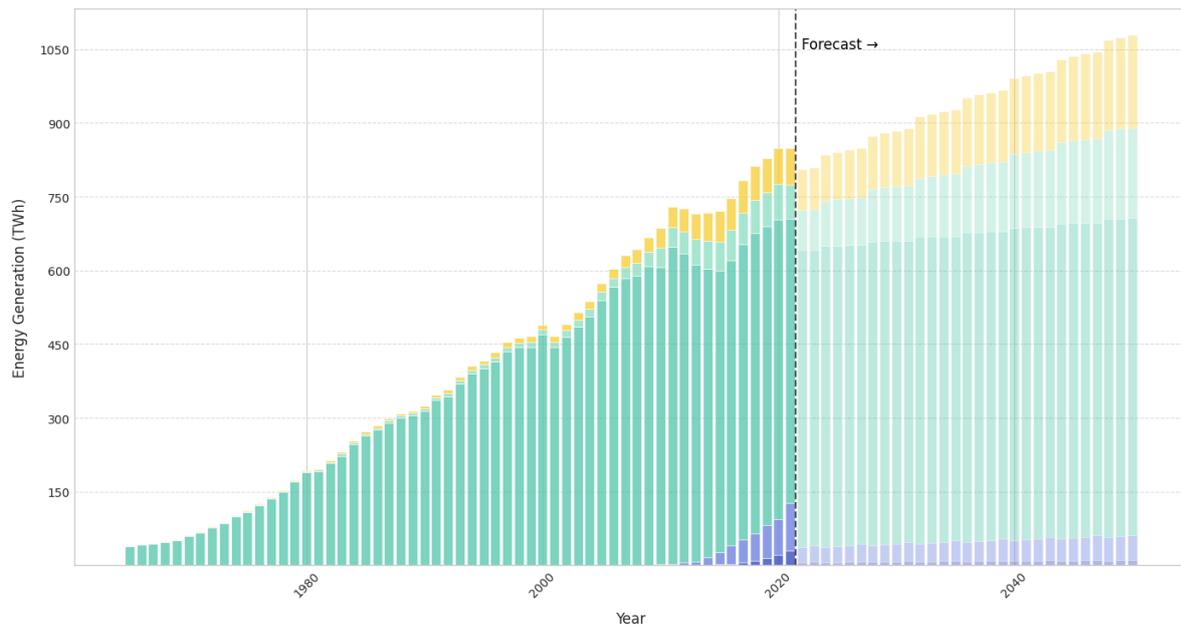

**Figure 5:** Renewable energy contribution in South America (Historical and forecast to 2050).

The stacked bar chart illustrates South America's historical and future contribution of various sources of renewable energy from decades ago to 2050. Forecasted with the assistance of a Prophet-based time series model, the projection highlights significant growth of renewable energy production, where solar and wind have the greatest expansions anticipated. A thick bar for historical data and transparency for forecasted output has been employed. A major point of change has been marked after 2023, suggesting a predicted multi-fold increase in total renewable production by 2050. Such a forecast gives policy makers and investors valuable insights regarding the changing world of energy as well as quick advancements towards clean energies.

## 4. Conclusion

This research provides relevant data on the future of renewable energy in South America in 2050, with a 300% increase in generation capacity and significant variation in sources of energy and countries. The research approach combining gradient boosting regression with Prophet Time series forecasting and geospatial mapping is a solid approach to recognizing pattern identification and opportunity detection by country. Key policy implications are the 2030-2040 investment window, diversification strategies across renewable resources, and interregional coordination to address country heterogeneity. These results are useful to policymakers, investors, and researchers for sustainable development in South America. While recognizing constraints in respect to long-term forecasting uncertainties, specifically technological upheavals, policy reforms, and geopolitical dynamics, the present study is a key addition to research on the trajectory of renewable energy in the region. Future research is required to explore scenario-based approaches, particularly to simulate these uncertainties and account for other factors such as investment flows, policy factors, and learning rates of the technology. Despite these constraints, the analysis provides an integrated foundation for strategic planning and international cooperation to enable South America's transition into a period of sustainable energy.